\documentclass{article}

 \usepackage[preprint]{neurips_2025}

\usepackage[utf8]{inputenc} 
\usepackage[T1]{fontenc}    
\usepackage{hyperref}       
\usepackage{url}            
\usepackage{booktabs}       
\usepackage{amsfonts}       
\usepackage{nicefrac}       
\usepackage{microtype}      
\usepackage{xcolor}         
\usepackage{graphicx}
\usepackage{float}
\title{ExpertAgent: Enhancing Personalized Education through Dynamic Planning and Retrieval-Augmented Long-Chain Reasoning}

\author{%
  Binrong Zhu \\
  Department of Computer Science \\
  San Francisco State University \\
  San Francisco, CA \\
  \texttt{bzhu2@sfsu.edu} \\
  \And
  Guiran Liu \\
  Department of Computer Science \\
  San Francisco State University \\
  San Francisco, CA \\
  \texttt{gliu@sfsu.edu} \\
  \And
  Nina Jiang\thanks{Corresponding author.} \\
  Department of Mechanical \& Industrial Engineering \\
  Louisiana State University \\
  Baton Rouge, LA \\
  \texttt{Nina.Jiang@lsu.edu} \\
}

\begin{document}

\maketitle

\begin{abstract}
The application of advanced generative artificial intelligence in education is often constrained by the lack of real-time adaptability, personalization, and reliability of the content. To address these challenges, we propose ExpertAgent—an intelligent agent framework designed for personalized education that provides reliable knowledge and enables highly adaptive learning experiences. 
Therefore, we developed ExpertAgent, an innovative learning agent that provides users with a proactive and personalized learning experience. ExpertAgent dynamic planning of the learning content and strategy based on a continuously updated student model. Therefore, overcoming the limitations of traditional static learning content to provide optimized teaching strategies and learning experience in real time. All instructional content is grounded in a validated curriculum repository, effectively reducing hallucination risks in large language models and improving reliability and trustworthiness.


\end{abstract}

\section{Introduction}

The rapid development of generative artificial intelligence (AI) and large language models (LLM) is profoundly transforming the landscape of education. They not only provide personalized support to learners, but also empower teachers and promote collaborative and simulation-based learning\cite{zhai2024transformingteachersrolesagencies}. 
Among these applications, AI agents have become an important form of generative AI in education, increasingly appearing in teaching practice and showing strong potential in diverse scenarios, ranging from general assistance to specialized tutoring in specific subjects.

However, current AI-in-education (AIED) systems face several critical challenges. Firstly, there is a serious lag in the integration of the state-of-the-art technologies into practical applications. Adoption in classrooms is constrained by teacher acceptance and existing pedagogical practices. In addition, limited model memory diminishes the coherence of long-term learning dialogues \cite{chu2025llmagentseducationadvances}. Besides, lack of effective personalization is hindering the wide adoption of a Generative AI-based educational assistant system. Most systems provide only superficial rules-based adaptations and lack emotional awareness, resulting in poor engagement of the learner \cite{milana2024artificial}. Above all, reliability and trustworthiness, which is a critical concern in the educational setting, is another major concern of both educators and students. Hallucinations, biased output, and potential overreliance are the most significant risks that must be addressed for the responsible implementation of AI in education \cite{mollick2024aiagentseducationsimulated}.

To address these challenges, we introduce ExpertAgent, an generative AI based agent framework designed for personalized and trustworthy educational assistant. 
Our contributions include:
The design and implementation of ExpertAgent, a novel framework that provides interactive and personalized learning experience for the users via integrating dynamic planning and RAG for reliable, adaptive educational content; 
We propose a student model that dynamically tracking the users' progress, providing feedback and planning learning contents;

\section{Literature Review}
\label{gen_inst}
AIED has expanded rapidly with the dual objectives of automating instruction and enabling personalized learning experiences. Existing systems are typically classified into pedagogical agents, which provide general support to teachers and learners, and domain-specific agents, which offer in-depth tutoring in subjects such as science, language, and law \cite{chu2025llmagentseducationadvances}. In this section, we conducted a literature review on the research across three key application domains, followed by enabling technical approaches and their limitations.

Artificial intelligence agents have shown great potential to simulate human tutors by providing individualized support. In particular, adaptive learning systems employ dynamic student modeling to track cognitive and emotional states, allowing real-time adjustments to instructional strategies and content \cite{chu2025llmagentseducationadvances}\cite{xu2024eduagentgenerativestudentagents}. In addition, AI can provide immediate feedback by analyzing student work in areas such as programming and writing, identifying errors, and offering corrective guidance to promote self-regulated learning \cite{wang2025impact}. Conversational partners such as ChatGPT further extend personalized support, simplifying complex concepts and providing 24/7 support \cite{milana2024artificial}. Despite these advances, most systems remain limited in their ability to deliver deep personalization that goes beyond surface-level adaptations.

Beyond tutoring, AI systems are increasingly helping teachers automate routine tasks and support instructional design. Generative models have been used to create lesson plans, instructional materials, and assessment strategies\cite{zhai2024transformingteachersrolesagencies}. Automated grading and feedback systems, such as the PROF framework, use reinforcement learning to deliver high-quality writing feedback while reducing the workload of teachers \cite{zhai2024transformingteachersrolesagencies}. Other systems recommend learning resources based on gaps in student knowledge or individual interests \cite{chu2025llmagentseducationadvances}. These applications highlight the capacity of AI to improve teacher efficiency, although they often prioritize productivity gains over pedagogical depth.

AI also enables the creation of low-cost and scalable simulation environments for practice and collaboration. Projects such as PitchQuest provide realistic role-play scenarios in which agents act as mentors, investors, and evaluators to train students in entrepreneurial pitching \cite{mollick2024aiagentseducationsimulated}. Similarly, teacher training simulations allow preservice teachers to rehearse classroom strategies in safe environments \cite{chu2025llmagentseducationadvances}. In collaborative learning contexts, AI agents can function as intelligent group members, offering technical support and facilitating team communication (Wang et al., 2025). Multi-agent systems further extend these possibilities by role-playing teachers, peers, and other actors, jointly fostering knowledge construction \cite{jiang2024aiagenteducationvon}. Although promising, these applications often remain limited to narrow domains or scripted scenarios.
    
The advances above are largely driven by three enabling techniques: retrieval-augmented generation (RAG), chain-of-thought (CoT) reasoning, and dynamic planning. RAG enhances factual reliability by incorporating output in external resources such as textbooks and validated curricula. However, its rigid pipeline makes it vulnerable to error propagation, where inaccurate retrievals directly contaminate generated responses, leading to confident but incorrect explanations. CoT improves reasoning transparency by requiring models to articulate intermediate steps, which is especially useful for multi-step problem solving and educational explanations. However, most CoT applications are limited to mathematical or logical reasoning tasks, offering limited adaptation to learners’ instructional trajectories. Dynamic planning, often studied within reinforcement learning, supports adaptive instruction by modeling student states and optimizing teaching strategies over time. However, defining appropriate educational rewards remains challenging, as overly simplistic signals (e.g., test scores) risk "reward hacking", emphasizing short-term performance at the expense of deep understanding\cite{chu2025llmagentseducationadvances}.

Despite their promise, these methods face persistent limitations. Systems suffer from retrieval latency that hampers interactivity, loss of long-term context that undermines dialogue coherence, error propagation across pipeline stages, insufficient multimodal capabilities for handling diverse educational materials, and fragile personalization that struggles to scale across heterogeneous learners\cite{chu2025llmagentseducationadvances}. More critically, prior work typically applies these methods in isolation: RAG is used mainly in open-domain QA with little curricular grounding; CoT is applied narrowly to reasoning tasks; and dynamic planning remains concentrated in robotics and gaming, with minimal adaptation to student modeling and learning trajectories\cite{zhai2024transformingteachersrolesagencies}\cite{chu2025llmagentseducationadvances}. These gaps underscore the need for integrated frameworks that combine reliability, reasoning, and adaptability in educational settings. Addressing this need, we propose ExpertAgent, a unified framework that leverages dynamic planning, RAG, and long-chain reasoning to deliver trustworthy and deeply personalized learning experiences.





\section{Methodology}
\begin{figure} [H]
  \centering
  \includegraphics[width=1\linewidth]{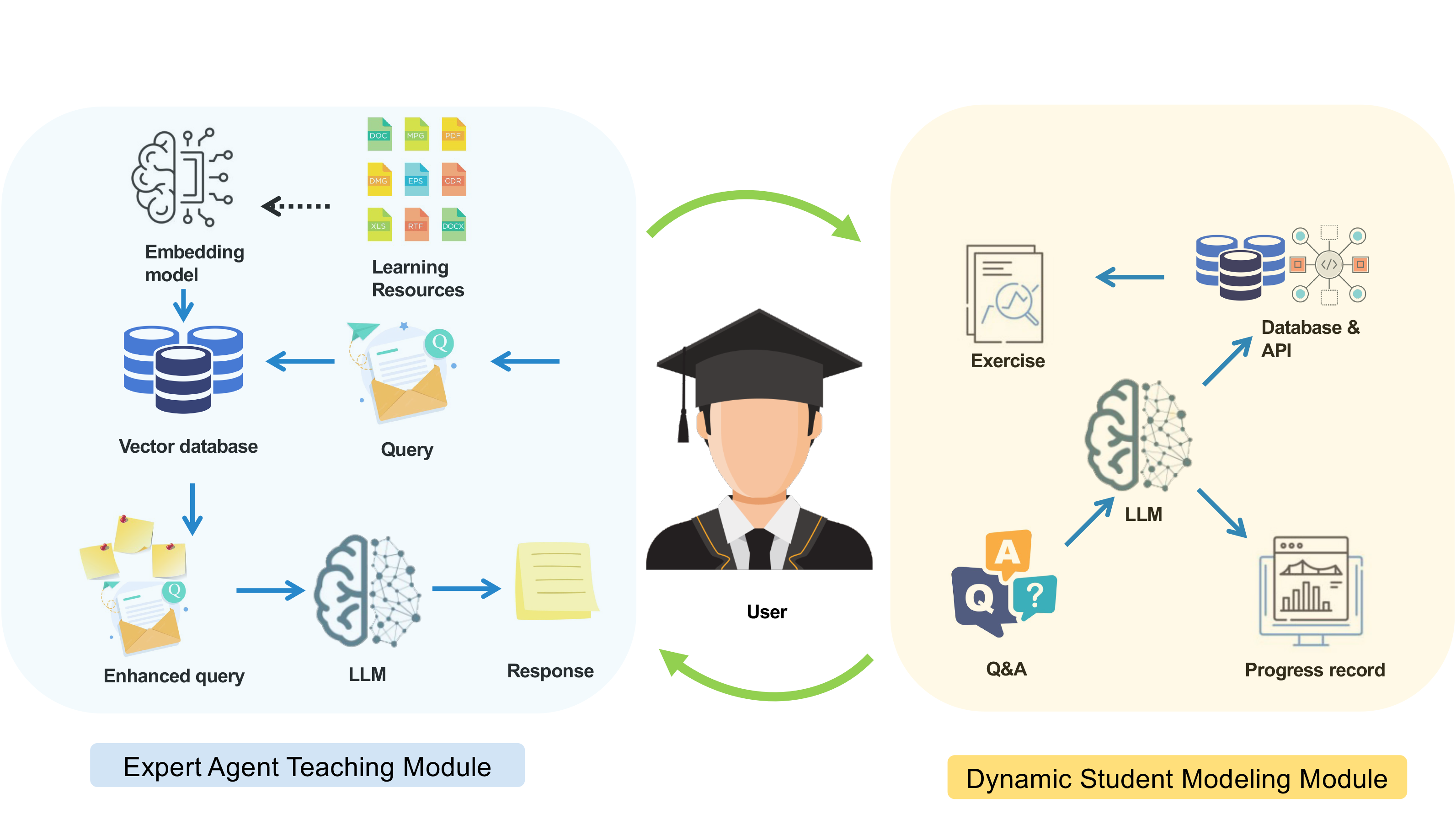}
  \caption{Overview of the ExpertAgent framework.}
  \label{fig:overview}
\end{figure}

To address those limitations, ExpertAgent integrates proactive teaching with a dynamically updated student model to provide a customized learning experience and adaptive personalized feedback and guidance based on users' progress and performance. Figure \ref{fig:overview} demonstrates the architecture of the ExpertAgent.

On the teaching side, ExpertAgent operates proactively. Instructional materials (such as textbooks and documents) are first segmented and embedded into a vector database. When a student issues a query, the system first retrieves the relevant context through RAG. The retrieved knowledge is then passed to the ExpertAgent module to enhance the prompt and a structured response is generated with CoT reasoning. This step not only provides correct answers, but also explains why a particular solution or concept applies and indicates the source of reference, thereby improving transparency and trustworthiness.

\label{headings}

On the student side, ExpertAgent maintains a structured student model that records progress across exercises, knowledge states, and performance trajectories. As shown in Figure 1, this model captures both the outcomes of the exercise and the evolving conceptual dependencies, functioning as a dynamic representation of the understanding of the learner. Through active learning and interactive feedback, the student model is continuously updated, enabling the system to adapt its teaching strategies in real time.

More importantly, the student model in ExpertAgent is not merely a passive recorder of performance, but an active driver of personalization. Our system updates the student model after each interaction, capturing both correct reasoning and misconceptions\cite{vsaric2024twenty}. These updates directly influence subsequent instructional planning, enabling the system to recommend targeted exercises, adjust task difficulty, and highlight weak or emerging concepts. Through this mechanism, the role of the learner shifts from being a passive recipient to becoming an active participant. Students can see their progress, receive tasks that suit their needs, and become more motivated and engaged in the learning process\cite{tschiatschek2022equityfairnessbayesianknowledge}.

The interaction between the two sides is achieved through the student's prompt and query input and the instructional feedback of the system. This bidirectional flow of information builds a teaching–learning feedback loop: the student submits a query or exercise response; ExpertAgent retrieves the relevant context and generates a pedagogically informed answer; the student model is updated; and the updated state, in turn, guides the next round of instructional planning.

This iterative cycle makes ExpertAgent not just an information provider, but an adaptive teacher capable of aligning instructional content with the evolving needs of the learner. The key innovation lies in the integration of RAG for factual reliability, CoT for reasoning transparency, and dynamic student modeling. Together, these three elements establish a foundation for trusting, interactive, and deeply personalized education.

\label{headings}
\begin{figure} [H]
  \centering
  \includegraphics[width=1\linewidth]{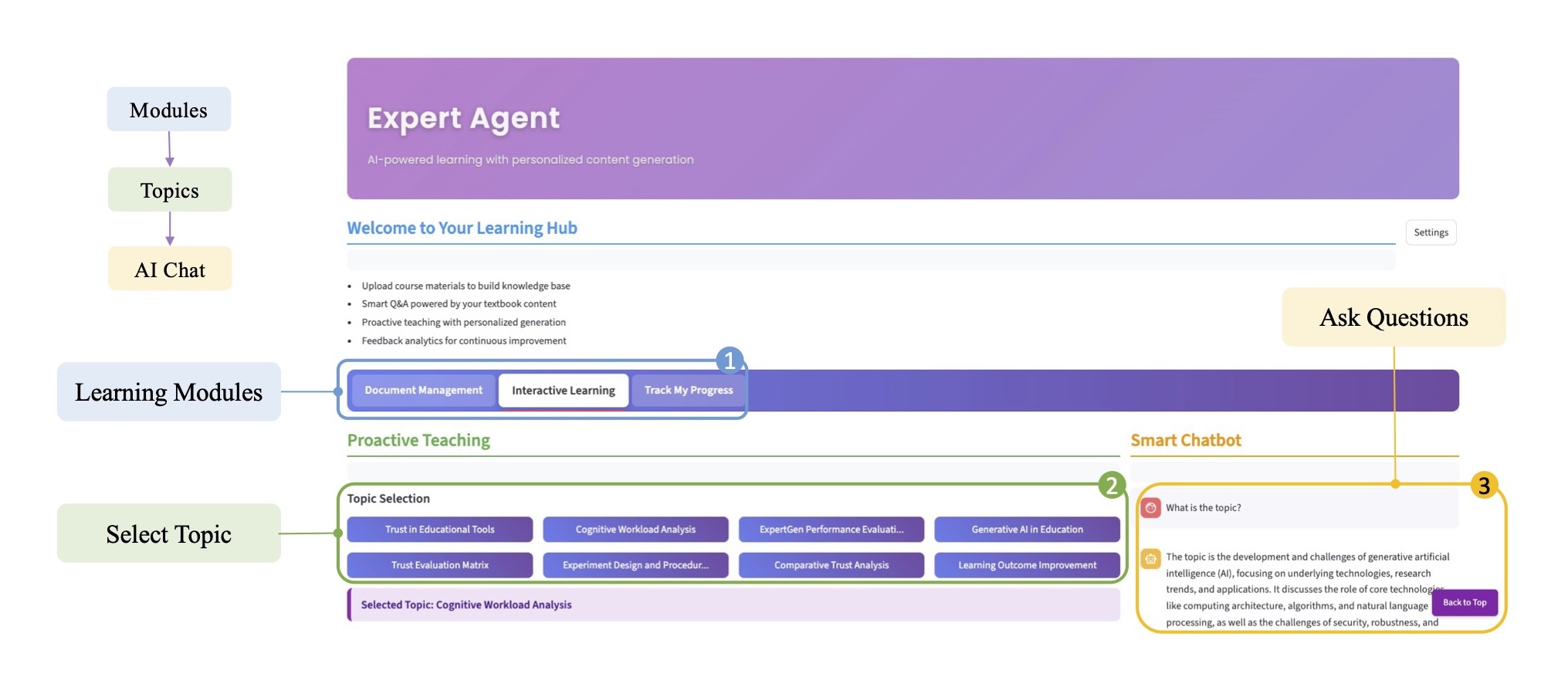}
  \caption[Functional flow diagram of the ExpertAgent system interface.]{
  Functional flow diagram of the ExpertAgent system interface. 
  The interface consists of three main components: 
  (1) Learning Modules, which allow switching between different learning modes; 
  (2) Select Topic, which enables learners to enter specific study topics; and 
  (3) Smart Chatbot, which provides interactive Q\&A and explanations based on RAG retrieval. 
  The overall learning flow follows a sequence of “Module Selection, Topic Selection, AI Q\&A,” 
  forming a comprehensive personalized learning pathway.}
  \label{fig:flow}
\end{figure}

\section{Results}
There are two major function of the ExpertAgent: (1) The Interactive Learning module to provide proactive teaching and personalized interaction to the users; (2) The "Track My Progress" module to provide interactive quizzes and knowledge maps to monitor learning performance and mastery; (3) The Document Management module, which processes multi-format materials and extracts key topics to support retrieval-augmented generation; (4) The Feedback Mechanism, which collects multi-level feedback from chat, teaching, and quizzes to evaluate system performance; (5) The Data Storage system, which records feedback and learning trajectories in structured formats (CSV and JSON) for analysis and continuous improvement; and (6) The Modular Architecture, which integrates the user interface, content generation, retrieval, and storage components to form an adaptive and scalable learning platform.

\begin{figure}[H]
  \centering
  \includegraphics[width=1\linewidth]{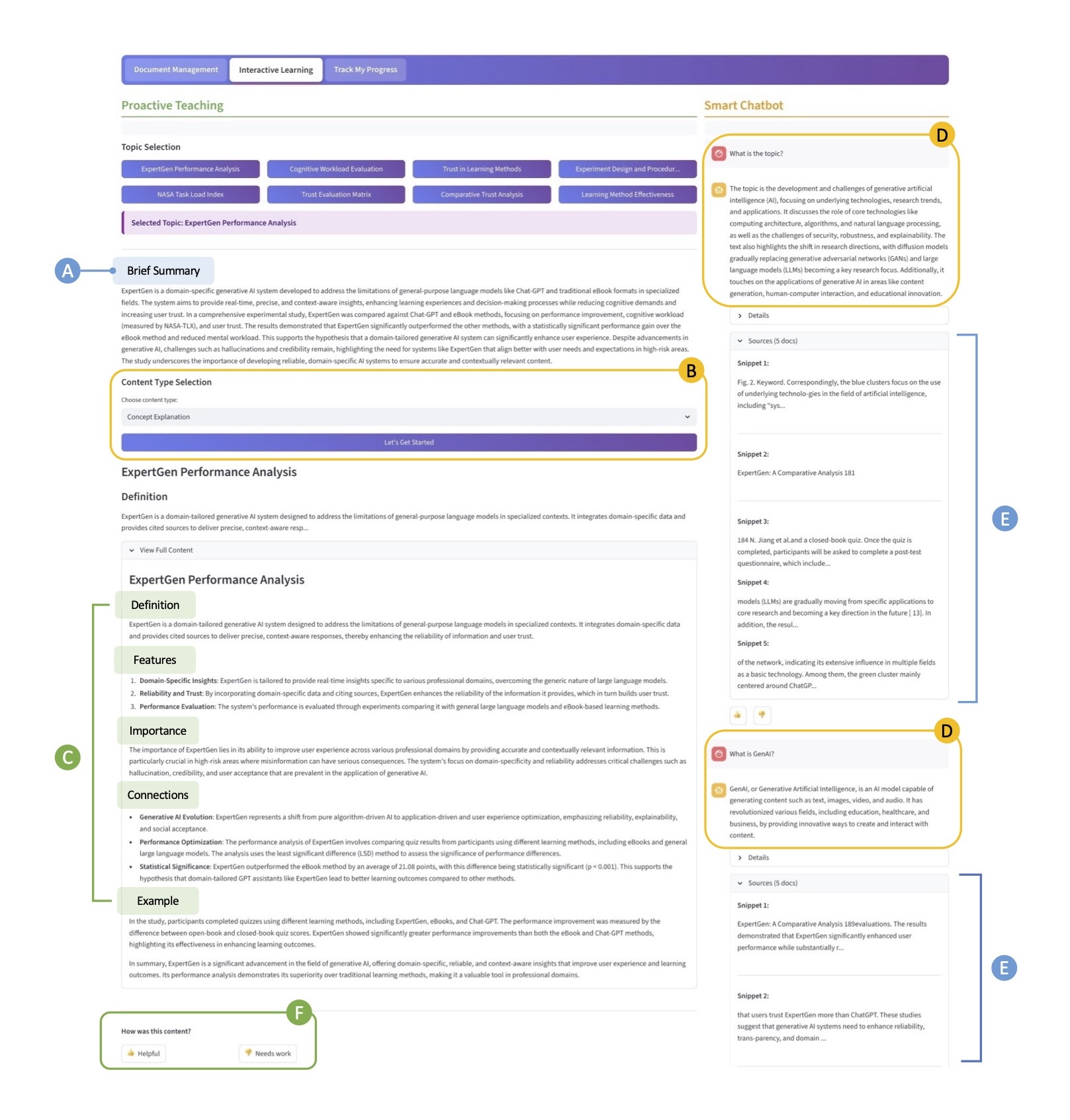}
  \caption[Structural diagram of the ExpertAgent teaching interface.]{
    Structural diagram of the ExpertAgent teaching interface. 
    The interface consists of six main components: 
    (A) Brief Summary, which provides a quick overview of the selected topic; 
    (B) Content Type Selection, which allows learners to choose different learning formats; 
    (C) Knowledge Details section, which provides structured and hierarchical content including definitions, 
    features, importance, connections, and examples, helping learners gain an in-depth understanding of the topic 
    and build systematic knowledge; 
    (D) AI Q\&A area, powered by RAG technology, which enables learners to interact with the intelligent assistant 
    and receive personalized explanations; 
    (E) Retrieved Snippets section, which supplies supporting evidence and contextual references from relevant documents; 
    and (F) Learning Feedback area, which collects learners’ evaluations of the usefulness of the generated content. 
    These components together form a complete loop from knowledge presentation to interactive questioning 
    and feedback-based improvement.}
  \label{fig:teaching}
\end{figure}

As illustrated in Figure \ref{fig:teaching}, demonstrate the structural design and key functions of the ExpertAgent teaching interface. The interface effectively supports personalized and interactive learning processes. The brief summary function helps learners quickly capture key information on a topic, thereby reducing the initial cognitive load. The content type selection module offers flexible learning pathways to cater to diverse needs. The Knowledge Details section presents structured and layered content through definitions, features, and examples, significantly enhancing the coherence and depth of the learning material. The AI Q\&A area, powered by RAG technology, delivers contextually relevant and personalized explanations, strengthening learning–system interaction. The retrieved snippets and feedback mechanism ensure traceability of information and continuous improvement. In general, these components form a complete learning loop that demonstrates positive effects in improving learning efficiency, improving knowledge mastery, and increasing learner satisfaction.

\begin{figure}[H]
  \centering
  \includegraphics[width=1\linewidth]{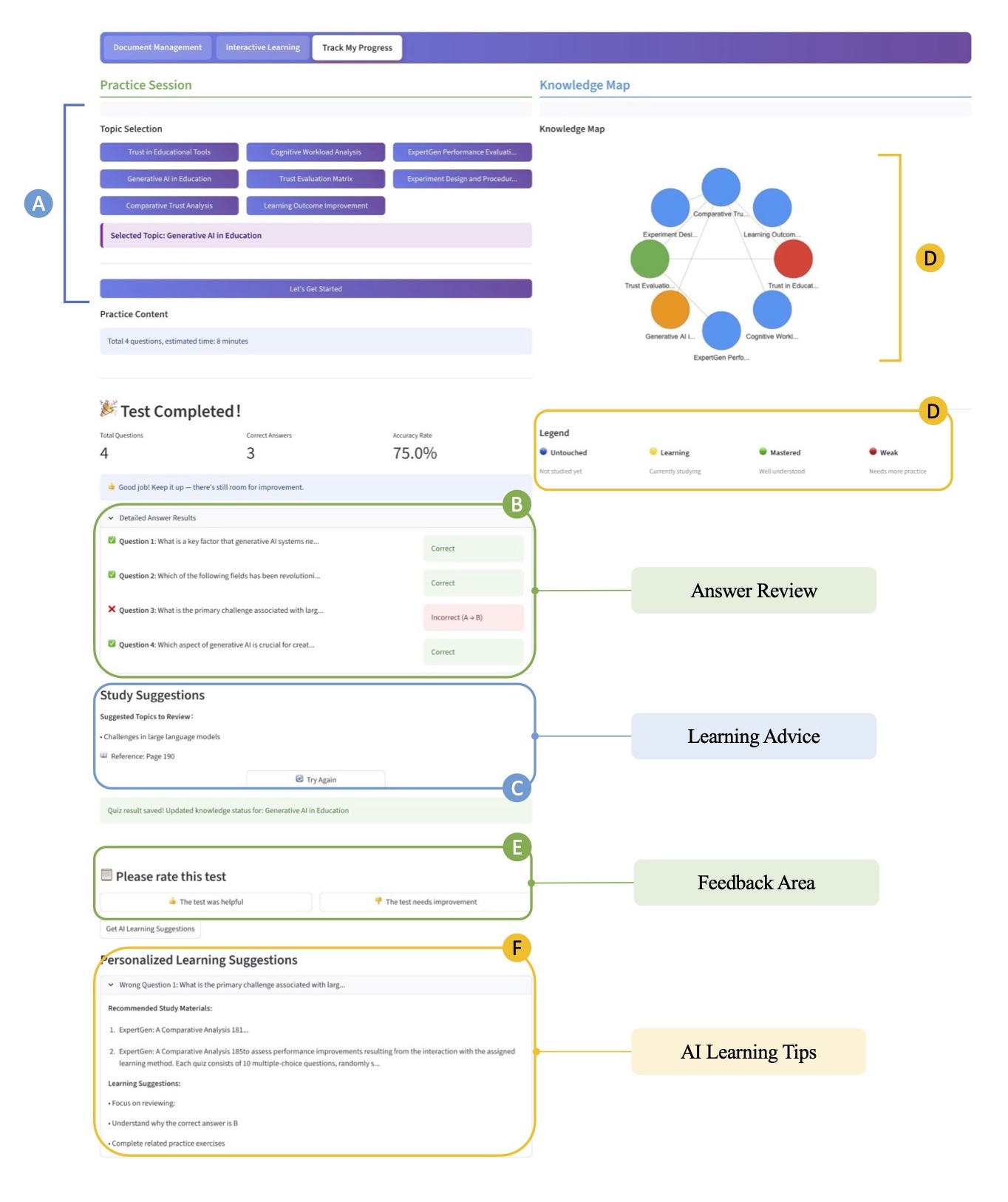}
  \caption[Practice and feedback interface of the ExpertAgent system.]{
    Practice and feedback interface of the ExpertAgent system. 
    The interface consists of five main components: 
    (A) Practice Session, which includes topic selection and practice arrangements; 
    (B) Answer Review, which displays learners’ correct and incorrect responses with instant feedback; 
    (C) Learning Advice, which provides review recommendations and reference materials based on test results; 
    (D) Knowledge Map, which visualizes the relationships between different learning topics and the learner’s mastery status 
    (blue = Untouched, yellow = Learning, green = Mastered, red = Weak); 
    (E) Feedback Area, which collects learners’ evaluations of the usefulness of the questions;
    and (F) AI Learning Tips, which generates personalized study guidance and improvement suggestions 
    based on the learner’s errors and weaknesses. 
    Overall, the interface supports a complete learning loop from test completion to result feedback, 
    learning advice, knowledge visualization, and personalized improvement.}
  \label{fig:practice}
\end{figure}

Figure \ref{fig:practice} shows the practice and feedback interface of the ExpertAgent system, showing that this module effectively supports learning assessment and personalized improvement. The practice session helps learners focus on core content through targeted topic selection and carefully arranged tasks. The answer review function provides instant feedback, enabling learners to identify and correct errors promptly. The Learning Advice module generates targeted review recommendations and reference materials based on test results, enhancing the relevance and efficiency of learning. The knowledge map visualizes the relationships between different topics and the mastery status of the learner, with blue indicating ‘Untouched’, yellow indicating ‘Learning’, green indicating ‘Mastered’, and red indicating ‘Weak’, thus offering an intuitive overview of the overall progress and weaknesses. The feedback area collects student evaluations of the usefulness and quality of the questions, providing evidence for system refinement. Finally, AI learning tips generate personalized guidance based on errors and weaknesses. In general, the interface forms a comprehensive learning loop, spanning from test completion and result feedback to personalized improvement, and demonstrates positive effects in supporting knowledge acquisition, enhancing learning efficiency, and facilitating continuous improvement.

\begin{figure}
  \centering
  \includegraphics[width=0.85\linewidth]{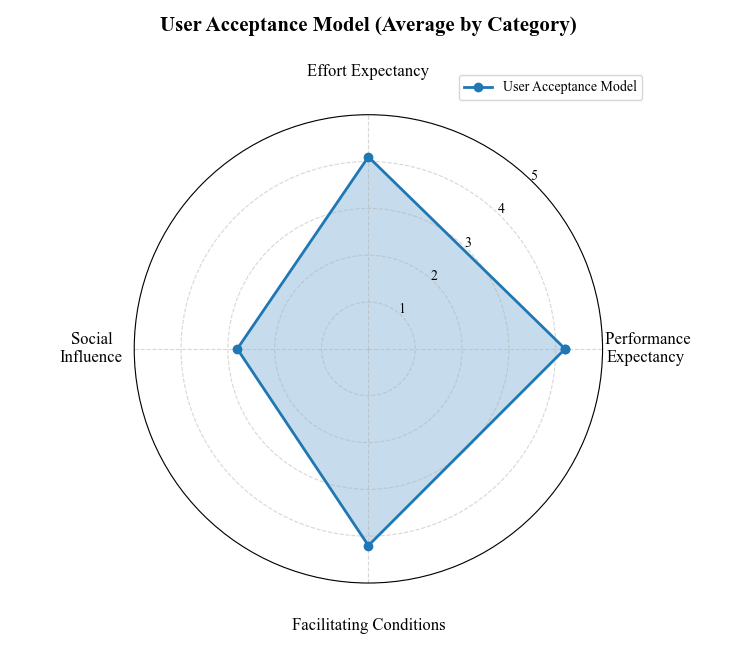}
  \caption{Average scores of the User Acceptance Model across four categories}
  \label{fig:radar}
\end{figure}

We conducted an internal test with volunteers to evaluate the user acceptance of ExpertAgent. 
The evaluation adopted the User Acceptance Model as the analytical framework, covering four categories:
performance expectancy, effort expectancy, social influence, and facilitation conditions. 
The participants rated each category on a Likert scale, and we calculated the average scores for each dimension
(see Figure~\ref{fig:radar}).

The results show that performance expectations (4.33), effort expectations (4.22) and facilitation conditions (4.22) achieved relatively high scores, 
indicating that users generally believed that the system could improve learning efficiency, was easy to use and provided adequate environmental and resource support. 
In contrast, Social Influence (2.78) was noticeably lower,
suggesting that external driving factors (e.g., peer recommendation or organizational support) 
remain limited for this system.

In general, the evaluation results show that ExpertAgent has been highly recognized by users
in terms of its core functionality and ease of use, while further improvements are needed in the dimension of social influence.

\section{Discussion}
The internal test results indicate that ExpertAgent provides interactive and personalized learning support. By integrating dynamic planning, retrieval-augmented generation, and long-chain reasoning, the system delivers clear explanations and professional guidance. Volunteer feedback suggests that ExpertAgent improves learning efficiency and helps students better understand complex concepts. The relatively high scores in performance expectancy, effort expectancy, and facilitating conditions reflect its usability and practical value, while the lower score on social influence highlights the need for stronger external adoption and peer-driven participation. 
Overall, the evaluation demonstrates that ExpertAgent promotes active learning and shows strong potential for deployment in real-world educational contexts.

\section{Conclusion and Future Work}
In summary, this paper developed the comprehensive framework for ExpertAgent, an interactive learning agent to foster active learning by providing feedback and guidance through the personalized student model and offering in-depth learning content and activities generated based on the retrieval of the validated learning materials. 
In the future, we aim to up-scale the experiment for diverse subjects, with different learning content and user groups. Through the longitude experiment, we will investigate the long-term impact of ExpertAgent on both educators and students.  
We also aim to strengthen mathematical and symbolic reasoning through reinforcement learning, allowing the system to assist in technical calculations, process analysis, and quantitative decision support. 
Ultimately, these advances will improve the ExpertAgent framework and have the potential to significantly enhance the learning experience and provide access to interactive and personalized instruction for all.

\bibliographystyle{unsrtnat}
\bibliography{references}

\end{document}